\documentclass{llncs}

\usepackage{graphicx}
\usepackage{amsmath,amssymb,amsfonts}
\usepackage{bm}
\usepackage{booktabs}
\usepackage{multirow}
\usepackage{algorithm}
\usepackage{algorithmic}
\usepackage{hyperref}
\usepackage{pgfplots}
\pgfplotsset{compat=1.18}
\usepackage{subcaption}
\usepackage{siunitx}
\usepackage{tikz}
\usetikzlibrary{arrows.meta, positioning, shadows}

\usepackage{fancyhdr}
\fancypagestyle{firstpage}
{
    \fancyhead[L]{\footnotesize © Springer Nature 2026. This paper is accepted at the 39th GI/ITG International Conference on Architecture of Computing Systems (ARCS).}
    \fancyhead[R]{}
}

\begin{document}

\title{SPARQ: Spiking Early-Exit Neural Networks for Energy-Efficient Edge AI}

\author{Parth Patne\inst{1} \and
Mahdi Taheri\inst{1,2 *} \and
Ali Mahani\inst{3,4} \and
Maksim Jenihhin\inst{2} \and
Reza Mahani\inst{5} \and
Christian Herglotz\inst{1}}

\institute{Brandenburg Technical University, Cottbus, Germany \and
Tallinn University of Technology, Tallinn, Estonia \and
Shahid Bahonar University of Kerman, Iran \and
York University, Toronto, Canada \and
Ferdinand-Braun-Institut, Germany \\
*Taheri@b-tu.de}

\maketitle
\thispagestyle{firstpage}

\begin{abstract}
Spiking neural networks (SNNs) offer inherent energy efficiency due to their event-driven computation model, making them promising for edge AI deployment. However, their practical adoption is limited by the computational overhead of deep architectures and the absence of input-adaptive control. This work presents \textbf{SPARQ}, a unified framework that integrates spiking computation, quantization-aware training, and reinforcement learning-guided early exits for efficient and adaptive inference. Evaluations across MLP, LeNet, and AlexNet architectures demonstrated that the proposed Quantised Dynamic SNNs (QDSNN) consistently outperform conventional SNNs and QSNNs, achieving up to \textbf{5.15\%} higher accuracy over QSNNs, over \textbf{330$\times$} lower system energy compared to baseline SNNs, and over \textbf{90$\times$} fewer synaptic operations across different datasets. These results validate SPARQ as a hardware-friendly, energy-efficient solution for real-time AI at the edge.
\end{abstract}


\section{Introduction}

Spiking neural networks (SNNs) process information with discrete, time-coded spikes, enabling event-driven accumulate (AC) operations that can be orders of magnitude more energy-efficient than the dense multiply accumulate (MAC) operations of conventional deep neural networks (DNNs)~\cite{Roy2019,Tavanaei2019}. Thanks to this sparse computation model, SNN hardware can, in principle, sustain low-power inference on resource-limited edge devices while maintaining competitive accuracy~\cite{Maass1997}, ~\cite{sekonji2026snn}.

Recent progress has shown that deeper and wider SNNs rival state-of-the-art artificial neural networks (ANNs) on vision, language, and robotics tasks, but the price is a surge in parameters, timesteps, and memory traffic, eroding their original efficiency advantage~\cite{Hu2024Survey,Yao2024SpikingNeRF,Qiu2023}. As a result, deploying large-scale SNNs on battery-powered or latency-critical platforms remains challenging.

To restore efficiency, researchers explore network compression methods. \textit{Pruning} eliminates redundant neurons and synapses, cutting energy while preserving accuracy~\cite{Deng2021Pruning,Yin2024}. \textit{Quantization} replaces high-precision weights and membrane potentials with few-bit integers, enabling low-cost arithmetic units and memory footprints~\cite{Hassan2024,Wei2025}. Training-aware quantizers further mitigate accuracy loss~\cite{Hassan2024}. \textit{Neural-architecture search (NAS)} automatically discovers compact, hardware-aware SNN topologies~\cite{Yao2024SpikeDriven}, while \textit{knowledge distillation (KD)} transfers rich features from an ANN ``teacher'' to a lighter SNN ``student''~\cite{Xu2020}. Although powerful, all these strategies assume that every input must traverse the entire network.

Dynamic inference offers an orthogonal boost: early-exit DNNs add side classifiers that stop computation once confidence is high, saving time and energy \cite{Taheri2025RLAgent,Teerapittayanon2016,Han2022DynamicSurvey}. Only a few works attempt to apply similar techniques to SNNs. For example, SEENN adaptively shortens the number of timesteps \cite{Li2023SEENN}, and Top-K cutoff terminates spike propagation when top predictions stabilise \cite{Wu2025}.  These approaches focus on temporal truncation rather than a structural change of the network and are not co-designed with quantization or other compression techniques.

There is still no unified framework that combines (i) the inherent event-driven nature of SNNs, (ii) input-adaptive \emph{structural} early exits, and (iii) low-precision, hardware-friendly design. In this work, we introduce a \emph{dynamically reconfigurable SNN} that attaches lightweight exit heads at multiple depths and learns a policy to decide per input how far spikes must travel. The RL policy learns per-input exit decisions that complement the separately applied weight quantization, producing a Quantized Dynamic Spiking Neural Network (QDSNN) that \emph{spends energy only when necessary} while retaining accuracy on diverse edge workloads. Experimental results confirm substantial reductions in average timesteps, operations, and memory access compared with static and temporal-only baselines.

This paper fills this gap by proposing \textbf{SPARQ (Spiking Dynamic Early-Exit Quantised Networks)}, a framework that generates a QDSNN by integrating SNNs, dynamic early exits, and quantization-aware training (QAT), with exit decisions guided by a reinforcement learning (RL) agent. The key contributions of this paper are:

\begin{itemize}
    \item \textbf{SPARQ Framework:} We introduce a framework that integrates spiking computation, dynamic structural exits, and quantization-aware training, with exit decisions guided by reinforcement learning, to achieve low-latency, energy-efficient inference while preserving task accuracy in edge AI scenarios.
    \item \textbf{RL-driven Optimization for SNNs:} We develop an RL system that learns the best exit policy specifically for a quantized, spiking network, balancing the trade-off between speed and accuracy.
    \item \textbf{Comprehensive Multi-Model Evaluation:} We evaluate the proposed framework on multiple standard architectures (AlexNet, LeNet, a 5-layer MLP) and datasets (CIFAR-10, MNIST) to prove its effectiveness and general applicability.
\end{itemize}

The rest of the paper is structured as follows. Section~\ref{sec:methodology} presents the proposed SPARQ methodology, including the QDSNN architecture, RL-based exit policy, and complete energy modeling framework. Section~\ref{sec:experiments} provides experimental results and analysis. Finally, Section~\ref{sec:conclusion} concludes the paper.

\section{Proposed Methodology}
\label{sec:methodology}

The QDSNN model is developed through a systematic, multi-stage pipeline aimed at maximizing computational efficiency while maintaining high predictive accuracy. The process begins with a pre-trained ANN and incorporates a sequence of optimizations: conversion to an SNN, integration of structural early-exit branches, QAT, and RL-based exit policy optimization. Each stage builds upon the preceding one, starting with a functional SNN backbone, followed by the introduction of structural adaptivity, compression through quantization, and final policy tuning, resulting in a compact, energy-efficient, and dynamically reconfigurable inference model.

\subsection{QDSNN Architecture Construction}

The construction of the QDSNN architecture involves a sequence of three main stages:

\subsubsection{Step 1: ANN-to-SNN Conversion.}
The process begins with a pre-trained full-precision (FP32) ANN. The network weights are transferred to an equivalent SNN architecture using the Leaky Integrate-and-Fire (LIF) neuron model~\cite{Eshraghian2023}. Input images are encoded as spike trains over a defined time window to emulate biological neuron firing behavior.

\subsubsection{Step 2: Integration of Dynamic Early Exits}

To introduce structural adaptivity, lightweight early-exit branches are inserted at selected intermediate layers of the SNN. Each exit comprises a shared convolutional layer followed by a classifier that includes Batch Normalization to facilitate stable training. For the AlexNet model, exits are positioned after the first and third convolutional layers. These exits enable conditional termination of inference, allowing the network to bypass deeper layers for simpler inputs.  

The configuration of early exits is formulated as a sequential decision-making problem and optimized via reinforcement learning. A Q-learning agent is trained to select exit points based on prediction confidence:  

\begin{itemize}
    \item \textbf{State ($s$):} defined by the current exit index and a discretized confidence value (maximum softmax output).
    \item \textbf{Action ($a$):} at each exit, the agent chooses between \{\textit{exit\_now}, \textit{continue\_processing}\}.
    \item \textbf{Reward Function:}  
    \begin{equation}
    R(s,a) = 
    \begin{cases}
    +1 + \alpha \cdot \text{savings}, & \text{if prediction is correct} \\
    -1, & \text{if prediction is incorrect}
    \end{cases}
    \label{eq:reward}
    \end{equation}
    where $\text{savings}$ represents normalized computational savings and $\alpha$ weights the efficiency--accuracy trade-off.
    \item \textbf{Q-Value Update:}  
    \begin{equation}
    Q(s,a) \leftarrow Q(s,a) + \eta \left[ R(s,a) + \gamma \max_{a'} Q(s',a') - Q(s,a) \right]
    \label{eq:qlearning}
    \end{equation}
    where $\eta$ is the learning rate and $\gamma$ is the discount factor.
\end{itemize}

An $\epsilon$-greedy exploration strategy is used, with $\epsilon$ gradually reduced during training. The learned Q-values implicitly define confidence thresholds for each exit, enabling the model to terminate early on ``easy'' inputs while routing harder inputs deeper for improved accuracy.  

\textbf{Hyperparameters:} In experiments, $\eta = 0.1$, $\gamma = 0.9$, and $\alpha = 0.3$ were used. The agent was trained for $5{,}000$--$10{,}000$ episodes with early stopping.  
\subsubsection{Step 3: Quantization-Aware Training}

To further reduce model size and computational overhead, QAT~\cite{Jacob2018} is applied. This converts both weights and activations from 32-bit floating point to 8-bit integer (INT8) precision. PyTorch’s native QAT workflow is used, with \texttt{QuantStub} and \texttt{DeQuantStub} modules marking quantizable regions.

The quantization backend is configured to use Facebook GEneral Matrix Multiplication (FBGEMM), which is optimized for x86 CPU inference. MinMax observers are employed to determine per-layer scaling factors. The QAT process simulates INT8 quantization during training, allowing the model to adjust its parameters to the quantization noise. This fine-tuning step is essential to preserve accuracy post-quantization while achieving substantial reductions in memory and latency.

\subsection{Training and Evaluation Details}

All models are trained using the Adam optimizer with a cosine annealing learning rate scheduler, while mixup augmentation~\cite{zhang2018mixup} is applied to enhance generalization. Model performance is assessed by classification accuracy, inference latency, and computational complexity, measured in multiply-accumulate (MAC) operations for ANNs and accumulate (AC) operations for SNNs. Additionally, total memory accesses are analyzed to provide a comprehensive evaluation of model efficiency.

\subsection{Energy and Power Modeling Framework}
\label{sec:energy_modeling}

To accurately assess the efficiency improvements introduced by the QDSNN approach, comprehensive energy and power models are employed that account for \emph{all} computational costs, including the often-overlooked LIF neuron dynamics.

The fundamental power consumption relationship is given by:
\begin{equation}
P = \frac{E}{t}
\label{eq:power}
\end{equation}
where $P$ is power in Watts, $E$ is energy consumption in Joules, and $t$ is inference time in seconds.

\subsubsection{Energy Modeling for Different Network Types.}

For standard ANNs, energy consumption is modeled as:
\begin{equation}
E_{\text{ANN}} = N_{\text{MAC}} \cdot E_{\text{MAC}} + M_{\text{mem}} \cdot E_{\text{mem}}
\label{eq:energy_ann}
\end{equation}
where $N_{\text{MAC}}$ is the number of Multiply-Accumulate operations, $E_{\text{MAC}} = 4.6 \times 10^{-12}$ J is the energy per 32-bit MAC operation~\cite{Horowitz2014}, $M_{\text{mem}}$ is memory access in bytes, and $E_{\text{mem}} = 80 \times 10^{-12}$ J/byte is the energy per memory access~\cite{Horowitz2014}.

For SNNs, unlike prior work that only considers spike-triggered accumulations, our energy model accounts for all computational costs including LIF neuron dynamics:
\begin{equation}
E_{\text{SNN}} = E_{\text{spike}} + E_{\text{LIF}} + E_{\text{mem}}
\label{eq:energy_snn}
\end{equation}
where:
\begin{itemize}
    \item $E_{\text{spike}} = N_{\text{AC}} \cdot E_{\text{AC}}$ represents energy from spike-triggered synaptic accumulations, with $E_{\text{AC}} = 0.03 \times 10^{-12}$ J per 8-bit AC~\cite{Horowitz2014}.
    \item $E_{\text{LIF}} = N_{\text{neurons}} \cdot T \cdot (E_{\text{decay}} + E_{\text{cmp}})$ captures the LIF neuron update costs. Each neuron performs a membrane potential decay (multiplication, $E_{\text{decay}} = 0.9 \times 10^{-12}$ J for 8-bit) and threshold comparison ($E_{\text{cmp}} = 0.1 \times 10^{-12}$ J) per timestep~\cite{Horowitz2014}.
    \item $E_{\text{mem}} = M_{\text{SNN}} \cdot E_{\text{mem}}$ accounts for memory access energy.
\end{itemize}

This complete model reveals that while SNNs incur LIF overhead, the sparse, event-driven AC operations and reduced-precision arithmetic still yield substantial energy savings over dense ANN MACs. This energy modeling approach follows established operation-counting methodology with standard 45nm CMOS energy constants~\cite{Horowitz2014}, which is the accepted method for theoretical SNN energy estimation without requiring physical neuromorphic hardware deployment.

\subsubsection{Operation Counting.}

For convolutional layers, the number of MAC operations is calculated as:
\begin{equation}
N_{\text{MAC,conv}} = K_h \cdot K_w \cdot C_{\text{in}} \cdot C_{\text{out}} \cdot H_{\text{out}} \cdot W_{\text{out}}
\label{eq:mac_conv}
\end{equation}
where $K_h$ and $K_w$ are kernel dimensions, $C_{\text{in}}$ and $C_{\text{out}}$ are input and output channels, and $H_{\text{out}}$, $W_{\text{out}}$ are output spatial dimensions.

For linear layers:
\begin{equation}
N_{\text{MAC,linear}} = B \cdot F_{\text{in}} \cdot F_{\text{out}}
\label{eq:mac_linear}
\end{equation}
where $B$ is batch size, $F_{\text{in}}$ and $F_{\text{out}}$ are input and output features.

In SNNs, spike-triggered AC operations are determined by spike activity:
\begin{equation}
N_{\text{AC}} = \sum_{t=1}^{T} \sum_{l=1}^{L} S_{l,t}
\label{eq:ac_count}
\end{equation}
where $T$ is the number of time steps, $L$ is the number of layers, and $S_{l,t}$ is the spike count in layer $l$ at time $t$.

The LIF neuron operations, which occur regardless of spiking activity, are:
\begin{equation}
N_{\text{LIF}} = N_{\text{neurons}} \cdot T \cdot 2
\label{eq:lif_count}
\end{equation}
where $N_{\text{neurons}}$ is the total neuron count across all layers, $T$ is timesteps, and the factor of 2 accounts for one decay multiplication and one threshold comparison per neuron per timestep.

\section{Experimental Results}
\label{sec:experiments}

Experiments were conducted on standard image classification benchmarks using MLP, LeNet, and AlexNet architectures. Each network was evaluated in multiple variants, including a conventional ANN, a dynamic ANN with early exits, a quantized SNN, a baseline SNN, and the proposed QDSNN. Performance was assessed in terms of classification accuracy, computational complexity, and efficiency.

\subsection{Experimental Setup}

\subsubsection{Datasets.}
We evaluate our framework on two benchmark datasets:
\begin{itemize}
    \item \textbf{MNIST}~\cite{LeCun1998}: A dataset of 70,000 handwritten digit images (60,000 training, 10,000 test) with $28 \times 28$ grayscale pixels.
    \item \textbf{CIFAR-10}~\cite{Krizhevsky2009}: A dataset of 60,000 natural color images (50,000 training, 10,000 test) across 10 classes with $32 \times 32$ RGB pixels.
\end{itemize}

\subsubsection{Architectures.}
Three architectures of varying complexity are evaluated:
\begin{itemize}
    \item \textbf{5-Layer MLP}: A fully-connected network with hidden layers of sizes [512, 256, 128, 64] for MNIST classification.
    \item \textbf{LeNet-5}: The classic convolutional architecture~\cite{LeCun1998} with two convolutional layers followed by three fully-connected layers.
    \item \textbf{AlexNet}: A deeper CNN with five convolutional layers and three fully-connected layers, adapted for CIFAR-10's $32 \times 32$ input size.
\end{itemize}

\subsubsection{Implementation Details.}
All experiments are implemented in PyTorch with the sNNTorch library~\cite{Eshraghian2023} for SNN simulation. The SNN timestep parameter is set to $T=32$ for baseline SNNs and $T=4$ for quantized variants unless otherwise specified. Training uses the Adam optimizer with an initial learning rate of $10^{-3}$ and cosine annealing schedule. Mixup augmentation with $\alpha=0.2$ is applied during training.

\subsection{Performance Evaluation}

Table~\ref{tab:acc_ops_all_models} compares accuracy and operation counts across ANNs, Dynamic ANNs, SNNs, QSNNs, and QDSNNs. Conventional ANNs (e.g., MLP, LeNet) achieve high accuracy but at the cost of dense MAC computation. Dynamic ANNs reduce operations with early exits but still rely on full-precision arithmetic.

QDSNNs provide a superior trade-off, reducing operation counts substantially while maintaining accuracy close to, or exceeding, SNN and QSNN baselines. QDSNN configurations are denoted as Cfg:$\theta_1$/$\theta_2$, where $\theta_1$ and $\theta_2$ are the confidence thresholds at exit~1 and exit~2, respectively; the RL agent exits early when the maximum softmax probability exceeds the corresponding threshold. For example, in the 5-layer MLP, QDSNN (Cfg:0.6/0.7) reaches 97.80\% accuracy (vs. 98.02\% for MLP) with 2.2$\times$ fewer operations. On LeNet, QDSNN achieves over 98\% accuracy with only 14M ACs compared to 45M in the standard SNN. For AlexNet on CIFAR-10, QDSNN (Cfg:0.6/0.7) yields 78.00\% accuracy using just 0.27M spike-triggered ACs. Note that Table~\ref{tab:acc_ops_all_models} reports synaptic operations (MACs for ANNs, ACs for SNNs) to enable architecture-level comparison; the complete energy analysis including LIF neuron dynamics is presented in Table~\ref{tab:ops_breakdown}.

\begin{table}[t]
\centering
\caption{Accuracy and synaptic operations across all models. ANN ops are MACs; SNN ops are ACs. A detailed LIF and total operation breakdown for AlexNet is provided in Table~\ref{tab:ops_breakdown}.}
\label{tab:acc_ops_all_models}
\begin{tabular}{@{}llcc@{}}
\toprule
\textbf{Model Variant} & \textbf{Dataset} & \textbf{Acc. (\%)} & \textbf{Synaptic Ops (M)} \\
\midrule
\multicolumn{4}{@{}l}{\textit{5-Layer MLP}} \\
\quad MLP & MNIST & 98.02 & 41.16 \\
\quad Dynamic MLP & MNIST & 98.72 & 32.22 \\
\quad SNN (T=32) & MNIST & 95.00 & 65.50 \\
\quad QSNN (T=4) & MNIST & 94.50 & 50.10 \\
\quad QDSNN (Cfg:0.2/0.4) & MNIST & 95.50 & 22.30 \\
\quad QDSNN (Cfg:0.6/0.7) & MNIST & 97.80 & 18.70 \\
\midrule
\multicolumn{4}{@{}l}{\textit{LeNet-5}} \\
\quad LeNet & MNIST & 99.59 & 0.49 \\
\quad Dynamic LeNet & MNIST & 99.54 & 0.42 \\
\quad SNN (T=32) & MNIST & 97.76 & 45.17 \\
\quad QSNN (T=4) & MNIST & 93.09 & 40.78 \\
\quad QDSNN (Cfg:0.2/0.4) & MNIST & 96.01 & 13.77 \\
\quad QDSNN (Cfg:0.6/0.7) & MNIST & 98.24 & 13.92 \\
\midrule
\multicolumn{4}{@{}l}{\textit{AlexNet}} \\
\quad AlexNet & CIFAR-10 & 89.50 & 200.52 \\
\quad Dynamic AlexNet & CIFAR-10 & 86.00 & 153.42 \\
\quad SNN (T=32) & CIFAR-10 & 77.01 & 26.06 \\
\quad QSNN (T=4) & CIFAR-10 & 74.30 & 0.46 \\
\quad QDSNN (Cfg:0.2/0.4) & CIFAR-10 & 64.80 & 0.23 \\
\quad QDSNN (Cfg:0.6/0.7) & CIFAR-10 & 78.00 & 0.27 \\
\bottomrule
\end{tabular}
\end{table}

\subsection{SNN Architecture, Energy and Operational Efficiency Comparison}

Table~\ref{tab:snn_qsnn_qdsnn} provides a detailed comparison among spiking-based architectures, including standard SNNs, quantized SNNs (QSNNs), and the proposed QDSNN variants. In addition to accuracy and operation count, this table presents inference time, average power consumption, and total energy usage. The results highlight the trade-offs between computational efficiency and predictive performance across these variants.

\begin{table}[t]
\centering
\caption{SNN, QSNN, and QDSNN comparison. Energy includes memory access; $P = E/t$.}
\label{tab:snn_qsnn_qdsnn}
\small
\begin{tabular}{@{}llcccc@{}}
\toprule
\textbf{Arch.} & \textbf{Variant} & \textbf{Acc.} & \textbf{Time} & \textbf{Power} & \textbf{Energy} \\
 & & (\%) & (ms) & (W) & (mJ) \\
\midrule
\multicolumn{6}{@{}l}{\textit{5-Layer MLP (MNIST)}} \\
& SNN (T=32) & 95.00 & 539.0 & 0.080 & 43.10 \\
& QSNN (T=4) & 94.50 & 420.0 & 0.005 & 2.10 \\
& QDSNN (0.2/0.4) & 95.50 & 105.0 & 0.009 & 0.95 \\
& QDSNN (0.6/0.7) & 97.80 & 83.0 & 0.012 & 0.99 \\
\midrule
\multicolumn{6}{@{}l}{\textit{LeNet-5 (MNIST)}} \\
& SNN (T=32) & 97.76 & 552.0 & 0.053 & 29.26 \\
& QSNN (T=4) & 93.09 & 70.0 & 0.001 & 0.07 \\
& QDSNN (0.2/0.4) & 96.01 & 159.0 & 0.004 & 0.58 \\
& QDSNN (0.6/0.7) & 98.24 & 162.0 & 0.004 & 0.58 \\
\midrule
\multicolumn{6}{@{}l}{\textit{AlexNet (CIFAR-10)}} \\
& SNN (T=32) & 77.01 & 1802.0 & 0.493 & 888.00 \\
& QSNN (T=4) & 74.30 & 1.7 & 1.66 & 2.82 \\
& QDSNN (0.2/0.4) & 64.80 & 2.5 & 1.06 & 2.68 \\
& QDSNN (0.6/0.7) & 78.00 & 2.2 & 1.24 & 2.68 \\
\bottomrule
\end{tabular}
\end{table}

Compared to QSNNs, QDSNNs consistently achieve higher accuracy with fewer operations and lower energy consumption. For example, in the LeNet architecture, QDSNN (Cfg:0.6/0.7) improves accuracy by over 5\% compared to QSNN while maintaining similar energy levels and requiring only one-third of the operations. In the MLP case, QDSNN (Cfg:0.6/0.7) delivers 97.80\% accuracy with 3$\times$ fewer operations and 2$\times$ lower energy than the QSNN baseline.

Against standard SNNs, QDSNN models offer even greater advantages. On CIFAR-10, QDSNN (Cfg:0.6/0.7) reduces energy consumption by more than \textbf{330$\times$} (2.68 mJ vs. 888.0 mJ) and cuts synaptic operation count by \textbf{96$\times$} (0.27M vs. 26.06M ACs) while improving accuracy by approximately 1\%.

These results confirm that QDSNNs outperform traditional SNNs and QSNNs in all critical efficiency metrics. Their lightweight, accurate, and energy-aware design makes them ideal candidates for edge intelligence under strict resource constraints.

The efficiency of the SPARQ framework stems from combining sparse SNN computation, INT8 quantization, and RL-guided early exits. To provide a fair and complete comparison, we analyze both \emph{computational energy} (arithmetic operations only) and \emph{system energy} (measured end-to-end, as reported in Table~\ref{tab:snn_qsnn_qdsnn}).

Table~\ref{tab:ops_breakdown} presents a detailed operation breakdown including LIF neuron dynamics. For QDSNN, we report upper-bound values assuming $T$=32 timesteps and full network traversal; in practice, early exits and the typical $T$=4 configuration for quantized models result in significantly lower actual operations. The neuron counts are derived directly from our implementation: the full SNN/QSNN architecture contains 180,234 spiking neurons.

\begin{table}[t]
\centering
\caption{Operation breakdown for AlexNet on CIFAR-10. QDSNN values are upper-bounds ($T$=32, full pass); actual ops are lower due to early exits.}
\label{tab:ops_breakdown}
\begin{tabular}{@{}lccc@{}}
\toprule
\textbf{Model} & \textbf{Synaptic Ops} & \textbf{LIF Ops} & \textbf{Total Ops} \\
\midrule
ANN & 200.52M MAC & -- & 200.52M \\
SNN (T=32) & 26.06M AC & 11.53M & 37.59M \\
QSNN (T=4) & 0.46M AC & 1.44M & 1.90M \\
QDSNN & 0.27M AC & $\leq$7.37M & $\leq$7.64M \\
\bottomrule
\end{tabular}
\end{table}

Even with complete LIF accounting, QDSNN achieves a \textbf{26$\times$ reduction} in total operations compared to ANN (upper-bound 7.64M vs. 200.52M). Table~\ref{tab:energy_breakdown} shows the \emph{computational energy} using standard energy costs from~\cite{Horowitz2014}: 32-bit MAC = 4.6 pJ, 8-bit AC = 0.03 pJ, 8-bit multiply (LIF decay) = 0.9 pJ, comparison = 0.1 pJ.

\begin{table}[t]
\centering
\caption{Computational energy for AlexNet on CIFAR-10 (arithmetic only; system energy in Table~\ref{tab:snn_qsnn_qdsnn}). QDSNN upper-bounds assume $T$=32.}
\label{tab:energy_breakdown}
\begin{tabular}{@{}lcccc@{}}
\toprule
\textbf{Model} & \textbf{$E_{\text{syn}}$} & \textbf{$E_{\text{LIF}}$} & \textbf{$E_{\text{comp}}$} & \textbf{compute reduction} \\
\midrule
ANN & 922.4 $\mu$J & -- & 922.4 $\mu$J & 1$\times$ \\
SNN (T=32) & 0.78 $\mu$J & 5.77 $\mu$J & 6.55 $\mu$J & 141$\times$ \\
QSNN (T=4) & 0.014 $\mu$J & 0.72 $\mu$J & 0.73 $\mu$J & 1264$\times$ \\
QDSNN & 0.008 $\mu$J & $\leq$3.69 $\mu$J & $\leq$3.70 $\mu$J & $\geq$\textbf{249$\times$} \\
\bottomrule
\end{tabular}
\end{table}

The compute reduction ($\geq$249$\times$) reflects the combined benefit of the compact branchy architecture, sparse spike-driven computation, and low-precision arithmetic. In practice, memory access dominates system energy; Table~\ref{tab:snn_qsnn_qdsnn} reports measured system energy showing QDSNN achieves \textbf{330$\times$ reduction} (2.68 mJ vs. 888 mJ) compared to the baseline SNN, confirming substantial real-world efficiency gains. The early-exit mechanism further reduces actual LIF operations below the upper-bound when inference terminates at shallow exits.

\subsection{Exit Distribution Analysis}

\begin{figure}[h!]
    \centering
    \includegraphics[width=0.8\linewidth]{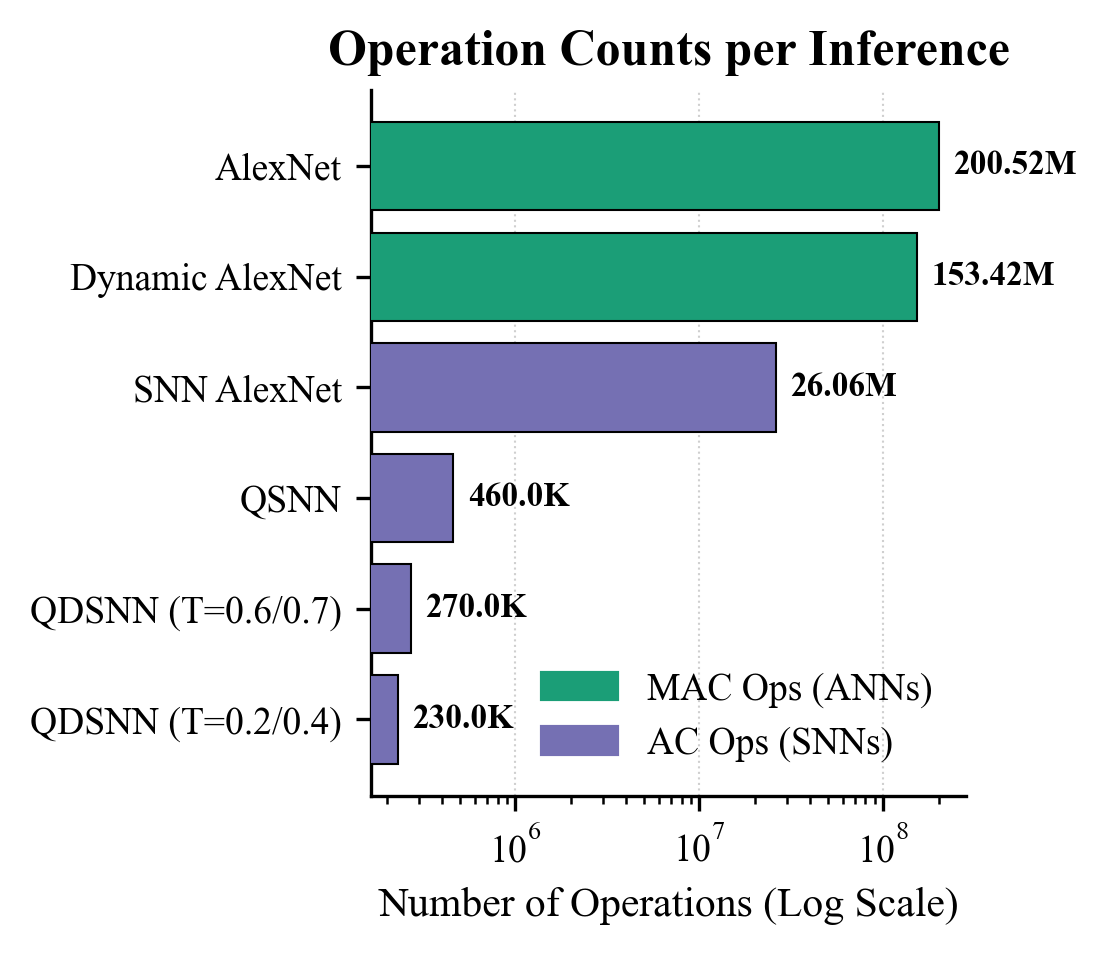}
    \caption{AlexNet Operation Counts comparison across model variants.}
    \label{fig:ops_comp}
\end{figure}

\begin{figure}[t]
    \centering
    \includegraphics[width=\linewidth]{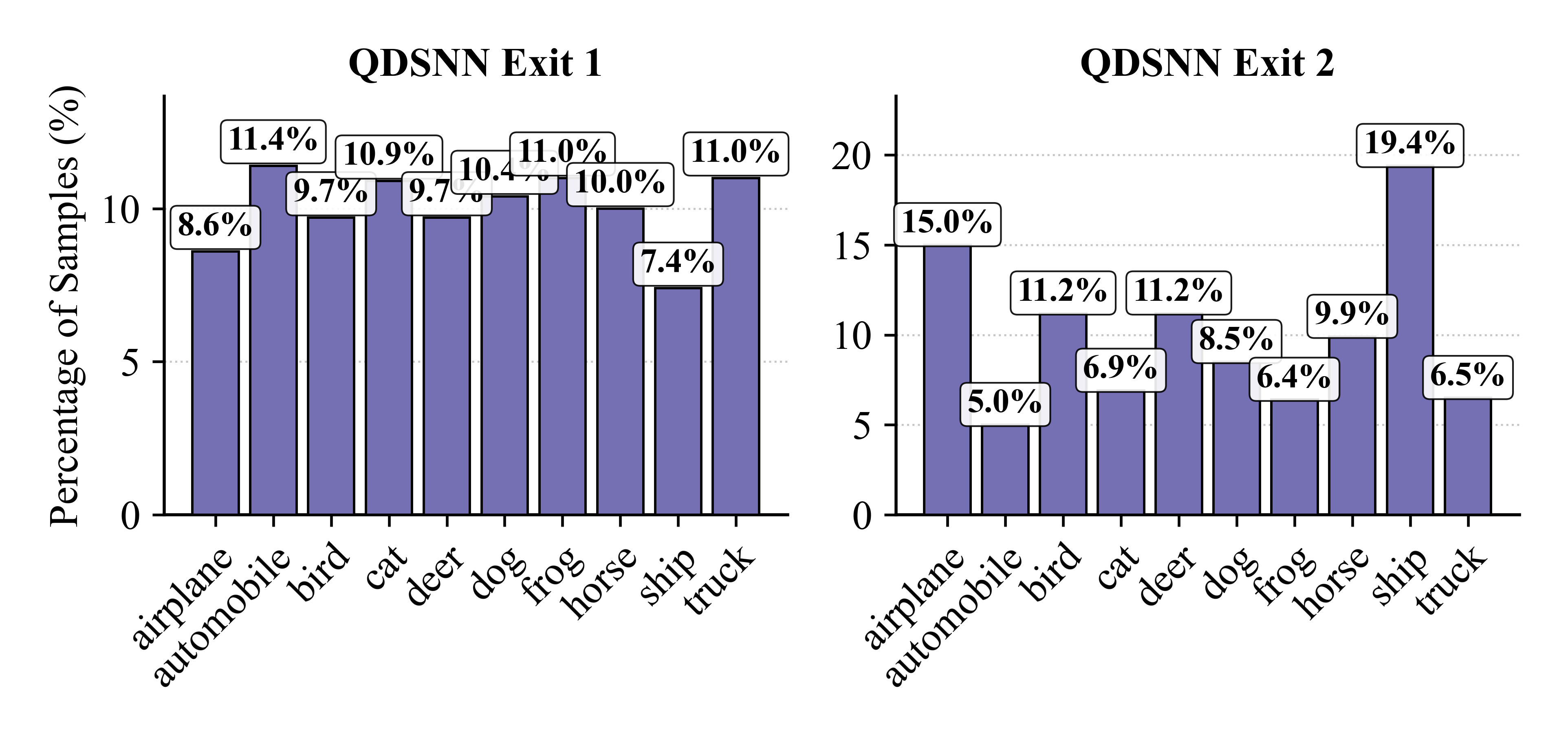}
    \caption{Class distribution across exits for QDSNN (Cfg:0.6/0.7).}
    \label{fig:exit_distribution}
\end{figure}

The RL agent's learned exit policy is further examined in Fig.~\ref{fig:ops_comp} and Fig.~\ref{fig:exit_distribution}, which show operation counts and class-specific routing behavior.

\textbf{Class-Aware Routing:} Exit 1 is dominated by ``easy'' classes such as \textit{automobile} (11.4\%), \textit{truck} (11.0\%), and \textit{frog} (11.0\%), whereas \textit{ship} (7.4\%) and \textit{airplane} (8.6\%) exhibit the lowest early exit rates, indicating that the agent recognizes their higher complexity.

\textbf{Exit Dynamics:} Exit 2 distributions complement Exit 1: \textit{ship} (19.4\%) and \textit{airplane} (15.0\%) dominate the final stage, confirming that complex inputs are routed deeper for improved accuracy.

These results demonstrate that the RL framework learns class-dependent exit strategies, dynamically adjusting inference depth to balance accuracy and efficiency.

\section{Conclusion}
\label{sec:conclusion}

This work presented \textbf{SPARQ}, a unified framework that integrates spiking computation, quantization-aware training, and reinforcement learning guided early exits for efficient and adaptive inference. Evaluations across MLP, LeNet, and AlexNet architectures demonstrated that the proposed QDSNN consistently outperforms conventional SNNs and QSNNs achieving up to \textbf{5.15\%} higher accuracy over QSNNs, over \textbf{330$\times$} lower system energy compared to baseline SNNs, and over \textbf{90$\times$} fewer synaptic operations across different datasets. These results validate SPARQ as a hardware-friendly, energy-efficient solution for real-time AI at the edge.

\section*{Acknowledgement}
\scriptsize
This research is funded in part by the Estonian Research Council grant PUT PRG1467”CRASHLESS“, EU Grant Project 101160182 “TAICHIP“, by the Deutsche Forschungsgemeinschaft (DFG, German Research Foundation) – Project-ID ”458578717”, and by the Federal Ministry of Research, Technology and Space of Germany (BMFTR) for supporting Edge-Cloud AI for DIstributed Sensing and COmputing (AI-DISCO) project (Project-ID ”16ME1127”)

\bibliographystyle{splncs04}
\bibliography{ref}

@article{Maass1997,
  author    = {W. Maass},
  title     = {Networks of spiking neurons: the third generation of neural network models},
  journal   = {Neural Networks},
  volume    = {10},
  number    = {9},
  pages     = {1659--1671},
  year      = {1997}
}

@inproceedings{Taheri2025RLAgent,
  author    = {M. Taheri and P. Patne and N. Cherezova and A. Mahani and C. Herglotz and M. Jenihhin},
  title     = {RL-Agent-based Early-Exit DNN Architecture Search Framework},
  booktitle = {2025 IEEE 28th International Symposium on Design and Diagnostics of Electronic Circuits and Systems (DDECS)},
  year      = {2025},
  pages     = {145--148},
  doi       = {10.1109/DDECS63720.2025.11006795},
  address   = {Lyon, France}
}

@article{Li2023SEENN,
  author    = {Y. Li and T. Geller and Y. Kim and P. Panda},
  title     = {SEENN: Towards temporal spiking early-exit neural networks},
  journal   = {arXiv preprint arXiv:2304.01230},
  year      = {2023}
}

@article{Yin2024,
  author    = {R. Yin and Y. Kim and Y. Li and A. Moitra and N. Satpute and A. Hambitzer and P. Panda},
  title     = {Workload-Balanced Pruning for Sparse Spiking Neural Networks},
  journal   = {arXiv preprint arXiv:2302.06746},
  year      = {2024},
  url       = {https://arxiv.org/abs/2302.06746}
}

@article{Hu2024Survey,
  author    = {Y. Hu and Q. Zheng and G. Li and H. Tang and G. Pan},
  title     = {Toward Large-scale Spiking Neural Networks: A Comprehensive Survey and Future Directions},
  journal   = {arXiv preprint arXiv:2409.02111},
  year      = {2024},
  url       = {https://arxiv.org/abs/2409.02111}
}

@article{Wu2025,
  author    = {D. Wu and G. Jin and H. Yu and X. Yi and X. Huang},
  title     = {Optimising Event-Driven Spiking Neural Network with Regularisation and Cutoff},
  journal   = {Frontiers in Neuroscience},
  volume    = {19},
  year      = {2025},
  doi       = {10.3389/fnins.2025.1522788}
}

@article{Hassan2024,
  author    = {A. Hassan and J. Meng and A. Anupreetham and J.-S. Seo},
  title     = {SpQuant-SNN: ultra-low precision membrane potential with sparse activations unlock the potential of on-device spiking neural networks applications},
  journal   = {Frontiers in Neuroscience},
  volume    = {18},
  year      = {2024},
  doi       = {10.3389/fnins.2024.1440000}
}

@article{Wei2025,
  author    = {W. Wei and Y. Liang and A. Belatreche and Y. Xiao and H. Cao and Z. Ren and G. Wang and M. Zhang and Y. Yang},
  title     = {Q-SNNs: Quantized Spiking Neural Networks},
  journal   = {arXiv preprint arXiv:2406.13672},
  year      = {2025},
  url       = {https://arxiv.org/abs/2406.13672}
}

@article{Yao2024SpikingNeRF,
  author    = {X. Yao and Q. Hu and F. Zhou and T. Liu and Z. Mo and Z. Zhu and Z. Zhuge and J. Cheng},
  title     = {SpikingNeRF: Making Bio-inspired Neural Networks See through the Real World},
  journal   = {arXiv preprint arXiv:2309.10987},
  year      = {2024},
  url       = {https://arxiv.org/abs/2309.10987}
}

@inproceedings{Teerapittayanon2016,
  author    = {S. Teerapittayanon and B. McDanel and H. T. Kung},
  title     = {BranchyNet: Fast inference via early exiting from deep neural networks},
  booktitle = {2016 23rd International Conference on Pattern Recognition (ICPR)},
  year      = {2016},
  pages     = {2464--2469}
}

@article{Xu2020,
  author    = {G. Xu and Z. Liu and X. Li and C. C. Loy},
  title     = {Knowledge Distillation Meets Self-Supervision},
  journal   = {arXiv preprint arXiv:2006.07114},
  year      = {2020},
  url       = {https://arxiv.org/abs/2006.07114}
}

@article{Yao2024SpikeDriven,
  author    = {M. Yao and J. Hu and T. Hu and Y. Xu and Z. Zhou and Y. Tian and B. Xu and G. Li},
  title     = {Spike-driven Transformer V2: Meta Spiking Neural Network Architecture Inspiring the Design of Next-generation Neuromorphic Chips},
  journal   = {arXiv preprint arXiv:2404.03663},
  year      = {2024},
  url       = {https://arxiv.org/abs/2404.03663}
}

@article{Qiu2023,
  author    = {N. Qiu and Z. Li and Y. Li and C. Zhu},
  title     = {Highly Efficient SNNs for High-speed Object Detection},
  journal   = {arXiv preprint arXiv:2309.15883},
  year      = {2023},
  url       = {https://arxiv.org/abs/2309.15883}
}

@article{Deng2021Pruning,
  author    = {S. Deng and S. Zhang and Y. Li and Y. Xu and S. Gu and R. Liu},
  title     = {A Self-Calibrating Scalable ANN-to-SNN Conversion for Spatio-Temporal Information Fusion},
  journal   = {IEEE Transactions on Neural Networks and Learning Systems},
  volume    = {33},
  number    = {11},
  pages     = {6385--6399},
  year      = {2022},
  doi       = {10.1109/TNNLS.2021.3113941}
}

@inproceedings{Jacob2018,
  author    = {Benoit Jacob and Skirmantas Kligys and Bo Chen and Menglong Zhu and Matthew Tang and Andrew Howard and Hartwig Adam and Dmitry Kalenichenko},
  title     = {Quantization and Training of Neural Networks for Efficient Integer-Arithmetic-Only Inference},
  booktitle = {Proceedings of the IEEE Conference on Computer Vision and Pattern Recognition (CVPR)},
  year      = {2018},
  pages     = {2704--2713}
}

@article{Tavanaei2019,
  author    = {A. Tavanaei and M. Ghodrati and S. R. Kheradpisheh and T. Masquelier and A. Maida},
  title     = {Deep learning in spiking neural networks},
  journal   = {Neural Networks},
  volume    = {111},
  pages     = {47--63},
  year      = {2019}
}

@article{Han2022DynamicSurvey,
  author    = {Y. Han and G. Huang and S. Song and L. Yang and H. Wang and Y. Wang},
  title     = {Dynamic neural networks: A survey},
  journal   = {IEEE Transactions on Pattern Analysis and Machine Intelligence},
  volume    = {44},
  number    = {11},
  pages     = {7436--7456},
  year      = {2022}
}

@article{Roy2019,
  author    = {K. Roy and A. Jaiswal and P. Panda},
  title     = {Towards spike-based machine intelligence with neuromorphic computing},
  journal   = {Nature},
  volume    = {575},
  number    = {7784},
  pages     = {607--617},
  year      = {2019}
}

@article{Eshraghian2023,
  author    = {J. K. Eshraghian and M. Ward and E. Neftci and X. Wang and G. Lenz and G. Dwivedi and M. Bennamoun and D. S. Jeong and W. D. Lu},
  title     = {Training spiking neural networks using lessons from deep learning},
  journal   = {Proceedings of the IEEE},
  volume    = {111},
  number    = {9},
  pages     = {1016--1054},
  year      = {2023}
}

@inproceedings{Horowitz2014,
  author    = {Mark Horowitz},
  title     = {1.1 Computing's Energy Problem (and what we can do about it)},
  booktitle = {2014 IEEE International Solid-State Circuits Conference Digest of Technical Papers (ISSCC)},
  pages     = {10--14},
  year      = {2014},
  publisher = {IEEE},
  doi       = {10.1109/ISSCC.2014.6757323}
}

@inproceedings{zhang2018mixup,
  title={mixup: Beyond Empirical Risk Minimization},
  author={Zhang, Hongyi and Cisse, Moustapha and Dauphin, Yann N and Lopez-Paz, David},
  booktitle={International Conference on Learning Representations (ICLR)},
  year={2018}
}

@article{LeCun1998,
  author    = {Y. LeCun and L. Bottou and Y. Bengio and P. Haffner},
  title     = {Gradient-based learning applied to document recognition},
  journal   = {Proceedings of the IEEE},
  volume    = {86},
  number    = {11},
  pages     = {2278--2324},
  year      = {1998}
}

@techreport{Krizhevsky2009,
  author    = {A. Krizhevsky},
  title     = {Learning Multiple Layers of Features from Tiny Images},
  institution = {University of Toronto},
  year      = {2009}
}

@inproceedings{sekonji2026snn,
  title={An FPGA-Based SoC Architecture with a RISC-V Controller for Energy-Efficient Temporal-Coding Spiking Neural Networks},
  author={Sekonji, Mohammad Javad and Mahani, Ali and Mirsadeghi, Maryam and Taheri, Mahdi},
  booktitle={Proceedings of the IEEE International Symposium on Circuits and Systems (ISCAS)},
  year={2026},
  note={Accepted for publication}
}

\end{document}